\documentclass[runningheads]{llncs}
\usepackage[T1]{fontenc}
\usepackage{microtype}
\usepackage{graphicx}
\usepackage{subfigure}
\usepackage{booktabs} 
\usepackage{amsmath}
\usepackage{amssymb}
\usepackage{booktabs}
\usepackage{multirow}
\usepackage{hyperref}
\usepackage{caption}
\usepackage{float} 
\usepackage{colortbl}
\usepackage{tablefootnote}
\usepackage{refcount}

\def\tablefootnotemark#1{\textsuperscript{\getrefnumber{#1}}}
\begin{document}
\title{Heavy-Tailed Regularization of Weight Matrices in Deep Neural Networks}
%
%
\author{Xuanzhe Xiao\inst{1}\and
Zeng Li\thanks{Corresponding author.}\inst{1} \and Chuanlong Xie\inst{2} \and Fengwei Zhou\inst{3}}
\authorrunning{X. Xiao et al.}
%
\institute{Southern University of Science and Technology, Shenzhen 518055, P.R. China 
\email{12132907@mail.sustech.edu.cn\\ liz9@sustech.edu.cn}
\and Beijing Normal University, Zhuhai 519087, P.R. China \\
\email{clxie@bnu.edu.cn}
\and Huawei Noah's Ark Lab, Hong Kong, P.R. China \\
\email{fzhou@connect.ust.hk}
}
\maketitle              
\begin{abstract}
Unraveling the reasons behind the remarkable success and exceptional generalization capabilities of deep neural networks presents a formidable challenge.
Recent insights from random matrix theory, specifically those concerning the spectral analysis of weight matrices in deep neural networks, offer valuable clues to address this issue. A key finding indicates that the generalization performance of a neural network is associated with the degree of heavy tails in the spectrum of its weight matrices.
To capitalize on this discovery, we introduce a novel regularization technique, termed \textbf{Heavy-Tailed Regularization}, which explicitly promotes a more heavy-tailed spectrum in the weight matrix through regularization.
Firstly, we employ the Weighted Alpha and Stable Rank as penalty terms, both of which are differentiable, enabling the direct calculation of their gradients. To circumvent over-regularization, we introduce two variations of the penalty function.
Then, adopting a Bayesian statistics perspective and leveraging knowledge from random matrices, we develop two novel heavy-tailed regularization methods, utilizing Power-law distribution and Fréchet distribution as priors for the global spectrum and maximum eigenvalues, respectively.
We empirically show that heavy-tailed regularization outperforms conventional regularization techniques in terms of generalization performance.

\end{abstract}
\section{Introduction}
Deep neural networks (DNN) have shown remarkable performance in recent years, achieving unprecedented success in various fields such as computer vision, natural language processing, and recommendation systems \cite{Chen_Zhao_Li_Huang_Ou_2019,Galassi_Lippi_Torroni_2020,he2016deep,Vaswani2017AttentionIA}.
However, there is still a lack of clear understanding of how neural networks generalize.
Efforts to construct a generalization framework for DNNs have incorporated various mathematical tools from conventional learning theory
\cite{bartlett1998almost,bartlett2017spectrally,bartlett2002rademacher,vapnik1994measuring}. 
Nevertheless, the majority of these approaches have been found to exhibit certain limitations.
For example, 
the VC dimension and Rademacher complexity have been deemed inadequate in offering a satisfactory explanation for the generalization performance of DNNs \cite{zhang2021understanding}. 
The uniform-convergence-based generalization bounds may fail to elucidate generalization in deep learning due to their vacuous generalization guarantee \cite{nagarajan2019uniform}.

One might consider that the weight matrices of a DNN serve as a representation of its generalization capabilities for the following reasons: 
from a theoretical standpoint, the parameters contained within the weight matrices are intricately connected to the model's output space, input data, optimization algorithm, etc;
Additionally, in practical scenarios, access to trained models often comes with limited information regarding training and testing data, which can be attributed to the highly compartmentalized nature of the industry.
Recently, Martin and Mahoney \cite{martin2021implicit} 
introduced a perspective grounded in random matrix theory (RMT) to elucidate the generalization behavior of deep neural networks.
They studied the empirical spectral distribution (ESD) of weight matrices in deep neural networks and observed a \textit{5+1 phase of regularization} -- throughout the training process, the ESDs of the weight matrices initially conform well to the Marchenko-Pastur (MP) law, gradually deviate from it, and ultimately approach a Heavy-Tailed (HT) distribution\cite{martin2019traditional,martin2021implicit}.
This regularization phenomenon is referred to as \textit{Implicit Self-Regularization}. 
Furthermore, this theory suggests that large, well-trained DNN architectures should exhibit Heavy-Tailed Self-Regularization, meaning that the spectra of their weight matrices can be well-fitted by a heavy-tailed distribution.
Building on Martin and Mahoney's work, Meng and Yao \cite{meng2021impact} discovered that the complexity of the classification problem could influence the weight matrices spectra of DNNs.
These theories offer a novel perspective for exploring the generalization of DNNs.

In addition to the aforementioned studies, several works have advocated the positive impact of heavy tails of weight matrices on the generalization of neural networks from the perspective of stochastic gradient descent (SGD).
Zhou et al. \cite{zhou2020towards} pointed out that the time required for both SGD and Adam to escape sharp minima is negatively related to the heavy-tailedness of gradient noise. 
They further explained that the superior generalization of SGD compared to Adam in deep learning is due to Adam's gradient calculation being smoothed by the exponential moving average, resulting in lighter gradient noise tails compared to SGD.
Hodgkinson et al. \cite{hodgkinson2021multiplicative} 
presented a similar finding, demonstrating that, within a stochastic optimization problem, multiplicative noise and heavy-tailed stationary behavior enhance the capacity for basin hopping during the exploratory phase of learning, in contrast to additive noise and light-tailed stationary behavior.
Simsekli et al. \cite{simsekli2020hausdorff} 
approximated the trajectories of SGD using a Feller process and derived a generalization bound controlled by the Hausdorff dimension, which is associated with the tail properties of the process.
Their results suggest that processes with heavier tails should achieve better generalization.
Barsbey et al. \cite{barsbey2021heavy} argued that the heavy-tailed behavior present in the weight matrices of a neural network contributes to network compressibility, thereby enhancing the network's generalization capabilities.
Taken together, these results suggest that the heavy tail of weight matrices is a fundamental factor for the improved generalization of DNNs under SGD.

An intuitive notion arising from these theories is that the presence of heavy tails of weight matrices during DNN training is crucial for achieving favorable generalization performance.
However, previous studies provide a limited understanding of how to enhance the heavy-tailed behavior in neural networks.
In this study,
our focus lies in regularizing DNNs to facilitate more rapid and pronounced heavy-tailed behavior.
To this end, we introduce an explicit regularization technique called \textbf{Heavy-Tailed Regularization}.
We empirically demonstrate that models trained with heavy-tailed regularization display superior generalization performance compared to those trained with conventional methods.

\subsubsection{Contribution of this paper.}

\begin{enumerate}
    \item We propose a regularization framework termed \textit{Heavy-Tailed Regularization}. 
    This proposal is motivated by prior research, which has shown that the heavy-tailed behavior of weight matrices in neural networks can improve their generalization capabilities.
    \item 
    We develop four distinct heavy-tailed regularization methods, including (a) Weighted Alpha regularization, (b) Stable Rank regularization, (c) Power-law Prior, and (d) Fr{\'e}chet Prior.
    The first two methods are inspired by existing complexity measures for neural networks, while the latter two are informed by insights from random matrix theory (RMT) and Bayesian statistics.
    \item We made comparison with conventional methods on widely used datasets including KMNIST and CIFAR10. Numerical experiments show that the heavy-tailed regularization approaches are efficient and outperform competing conventional regularization methods.
\end{enumerate}

\section{Heavy-Tailed Regularization}

\subsection{Definition}

Consider a DNN $f_{\mathbf{W}}:\mathcal X \rightarrow \mathcal{Y}$ with $L$ layers and with weight matrices of its fully connected layers $ \mathbf{W}=\left\{ \mathbf{W}_1,\mathbf{W}_2,\cdots, ,\mathbf{W}_L \right\} $ and data sample set $S=\left\{ \left( x_1,y_1 \right) ,\left( x_2,y_2 \right) \cdots ,\left( x_N,y_N \right) \right\}$ with sample size $N$. \par
Denote $l(f(x),y)$ the loss of example $(x,y) \in \mathcal X \times \mathcal Y$ under model $f_{\mathbf{W}}$. 
The optimization problem of the DNN can be viewed as a problem of minimizing the empirical risk with a penalty term:
\begin{equation}
\label{prmp}
    \min_{\mathbf{W}}\quad \mathcal L\left( x,y \right) =\frac 1N \sum_{i=1}^N{l\left( f\left( x_i \right) ,y_i \right)}+\lambda \sum_{l=1}^L{p_l\left( \mathbf{W}_l\right)},
\end{equation}
where $\lambda$ is a tuning parameter and  $p_l\left( \cdot \right)$ is a penalty function on the weight matrices. \par
Here, we propose a class of regularization methods called \textbf{Heavy-Tailed Regularization}, which refers to the regularization methods that are conducive to making the model's weight matrices more heavy-tailed. To achieve this goal, $p_l\left( \cdot \right)$ is supposed to be a complex measure of the model that reflects the degree of the heavy-tailed behavior of the model, and it decreases as the tails get heavier. \par
To describe the degree of heavy tails of the spectra of the weight matrices, it is critical to estimate the tail index $\alpha$. In statistics, 
estimating the tail index is a tricky issue. 
Denote the data points $\{x_i, 1\leq i\leq n\}$ and assume the data come from a heavy-tailed distribution with density function $p(x) \sim c x^{-\alpha}$, i.e., its p.d.f. is comparable with \textit{power-law $x^{-\alpha}$} as $x\rightarrow \infty$. 
A general method to estimate the tail index $\alpha$ is the Hill estimator (HE), which can be used for general power-law settings. If the data is sorted in increasing order, the HE can be written by:
\begin{equation}
\hat{\alpha}=1+\frac{k}{\left( \sum_{i=1}^k{\ln \frac{x_{n-i+1}}{x_{n-k}}} \right)},
\end{equation}
where $k$ is a tuning parameter. There is a trade-off depending on the value of $k$ between the bias and variance of the estimator. In this study, we use HE with $k=\frac{n}{2}$ for tail index estimation.

\subsection{Weighted Alpha Regularization}
Motivated by Martin and Mahoney's work \cite{martin2020heavy,martin2021post,martin2021predicting}, the \textit{Weighted Alpha} (also called \textit{AlphaHat}) is used in our regularization approach. In their theory, there is a strong linear correlation between the test accuracy and the Weighted Alpha of models. The Weighted Alpha is defined as:
\begin{equation}
\text{Weighted Alpha}(\mathbf{W})=\sum_{l=1}^L{\alpha _l\log \lambda _{\max ,l}},
\end{equation}
where $\alpha_l$ is the tail index of all the positive eigenvalues of $\mathbf{S}_l=\mathbf{W}^T _l\mathbf{W}_l$, and $\lambda _{\max,l}$ is the maximum eigenvalue of $\mathbf{S}_l$. 
In Martin and Mahoney's theory, only the performance of Weighted Alpha on different architectures of large-scale, pre-trained, state-of-the-art models was discussed. While we are interested in how this metric changes during the training process of DNNs. Here, we conducted some experiments and obtained evidence that Weighted Alpha is \textit{negatively correlated} with test accuracy. Thus, the penalty function $p_l(\cdot)$ can be written as
\begin{equation}
p_l\left( \mathbf{W}_l \right) =\alpha _l\cdot \log \lambda _{\max ,l}.
\end{equation}
\par
In fact, we do not need to penalize the weighted alpha throughout the training process. 
{Our goal is to impose a heavy-tailed perturbation in the stochastic optimization, which only requires us to activate the regularization in the early stages or intermittently. Otherwise, it will be over-regularized. 
On the other hand, for practical reasons, we can terminate the regularization at some point to avoid high computational costs.
Therefore, we provide two additional variants of the penalty function as follows:
\begin{enumerate}
    \item Decay Weighted Alpha: 
    \begin{equation}
        p_l\left( \mathbf{W}_l \right) =d\left( {{\lfloor e / m \rfloor}} \right) \cdot \alpha _l\cdot \log \lambda _{\max ,l},
    \end{equation}
    where $e$ is the current epoch number, $m$ is the frequency of decay, and $d(\cdot)$ is a decreasing function called the \textit{decay function}. The decay function is called \textit{power decay} when $d\left( x \right) =x^{-k} I _\{x^{-k}>t\}$ and called \textit{exponential decay} when $d\left( x \right) =\exp\left(-kx\right) I_\{\exp\left(-kx\right)>t\}$ for the hyperparameter $k$ and $t$. The adoption of this penalty function means that the regularization is activated only in the early epochs and becomes weaker with training.
    \item Lower Threshold Weighted Alpha:
    \begin{equation}
        p_l\left( \mathbf{W}_l \right) =\alpha _l\cdot \log \lambda _{\max ,l} \cdot I{\left\{ \sum_{l=1}^L{\alpha _l\cdot \log \lambda _{\max ,l}}\geqslant t \right\}},
    \end{equation}
    where $t$ is a hyperparameter, $I\left\{\cdot\right\}$ is the indicator function. The adoption of this penalty function means that the regularization is activated only when the Weighted Alpha is above a predetermined lower threshold $t$, i.e., the model falls short of the degree of 
    the heavy tail we expect.
\end{enumerate}
\subsection{Stable Rank Regularization}
Stable rank is a classical metric in deep learning, which is defined as
\begin{equation}
    \text{stable}\left( \mathbf{W}_l \right) =\frac{\left\| \mathbf{W}_l \right\|^2 _F}{\left\| \mathbf{W}_l \right\| ^2_2}.
\end{equation}
It has been verified that the stable rank decreases as the training of DNNs \cite{martin2021implicit}. Several recent studies \cite{bartlett2017spectrally,neyshabur2017pac} also shows that the generalization error can be upper bounded by $O\left( \prod_i{\left\| \mathbf{W}_i \right\| _{2}^{2}}\sum_i{\mathrm{stable}\left( \mathbf{W}_i \right)} \right)$, which implies that a smaller $\sum_i{\mathrm{stable}\left( \mathbf{W}_i \right)}$ leads to a smaller generalization error. In light of this, the penalty function for the stable rank regularization can be written as $p_l\left( \mathbf{W}_l \right) = \text{stable}\left( \mathbf{W}_l \right)$, and thus the optimization problem can be written as
\begin{equation}
\min_{\mathbf{W}} \quad \mathcal{L} \left( x,y \right) =\sum_{i=1}^N{l\left( f\left( x_i \right) ,y_i \right)}+\lambda \sum_{l=1}^L{\frac{\left\| \mathbf{W}_l \right\| ^2_F}{\left\| \mathbf{W}_l \right\| ^2_2}}
\end{equation}
Note that $||\mathbf W||_F^2$ is the sum of square singular values of $\mathbf{W}$ and $||\mathbf W||_2$ is the maximum singular value of $\mathbf W$. Recall the random matrix theory, when the matrix is heavy-tailed, the maximum eigenvalue is far off the global spectrum. Combined with Martin's 5+1 phase transition theory \cite{martin2021implicit}, a smaller stable rank of the weight matrix leads to stronger heavy-tailed self-regularization.

Similar to the weighted alpha regularization, in order to avoid over-regularization, we can also add decay and upper threshold to stable rank regularization as follows:
\begin{enumerate}
    \item Decay Stable Rank: 
    \begin{equation}
        p_l\left( \mathbf{W}_l \right) =d\left( {{\lfloor e / m \rfloor}} \right) \cdot \frac{\left\| \mathbf{W}_l \right\| ^2_F}{\left\| \mathbf{W}_l \right\| ^2_2}.
    \end{equation}
    \item Lower Threshold Stable Rank:
    \begin{equation}
        p_l\left( \mathbf{W}_l \right) =\frac{\left\| \mathbf{W}_l \right\| ^2_F}{\left\| \mathbf{W}_l \right\| ^2_2}\cdot I{\left\{ \sum_{l=1}^L{\frac{\left\| \mathbf{W}_l \right\| ^2_F}{\left\| \mathbf{W}_l \right\| ^2_2}}\geqslant t \right\}}.
    \end{equation}
\end{enumerate}

\subsection{Heavy Tailed Regularization from A Bayesian Perspective}

Here, we propose two heavy-tailed regularization methods from a Bayesian perspective. Let us view the deep neural network as a probabilistic model $P(\mathbf y |\mathbf x, \mathbf W)$, where $\mathbf{x} \in \mathcal X = \mathbb R ^p$ is the input and $\mathbf y \in \mathcal Y$ is the output probability assigned by the neural network. $\mathbf W=\{\mathbf W_1,\cdots, \mathbf W_L\}$ is the set of weight matrices of the neural network. Given a training sample set $S=\left\{ \left( x_1,y_1 \right) ,\left( x_2,y_2 \right) \cdots ,\left( x_N,y_N \right) \right\}$ with sample size $N$, a common method for estimating the weights $\mathbf W$ is the maximum likelihood estimation (MLE):
\begin{align}
\mathbf{W}^{\mathrm{MLE}}=\underset{\mathbf{W}}{\mathrm{arg}\max}\sum_{i=1}^N{\log P\left( y_i\left| x_i,\mathbf{W} \right. \right)}.
\end{align}
Specifically, for a multi-classification task, the probabilistic model is usually a multinomial distribution, and then the MLE can be written as
\begin{align}
\mathbf{W}^{\mathrm{MLE}}=\underset{\mathbf{W}}{\mathrm{arg}\max}\sum_{i=1}^N{y_i\log f_{\mathbf{W}}\left( x_i \right)}.
\end{align}
From a Bayesian perspective, if we want to introduce the heavy-tailed regularization upon the model, we can assign a heavy-tailed prior upon the probabilistic model and then find the maximum a posteriori (MAP) rather than MLE:
\begin{equation}
\label{eq:bayes}
\mathbf{W}^{\mathrm{MAP}}=\underset{\mathbf{W}}{\mathrm{arg}\max}\log P\left( \mathbf{y}\left| \mathbf{x},\mathbf{W} \right. \right) +\log P\left( \mathbf{W} \right). 
\end{equation}
Thus, it is important to choose a reasonable prior $P(\mathbf W)$ for the weights which can make the weights more heavy-tailed. \par
Recall the Random Matrix Theory for the heavy-tailed matrices, if a random matrix is heavy-tailed, its limiting spectral distribution (LSD) is supposed to be power-law \cite{davis2016extreme,davis2016asymptotic,davis2014limit} and its largest eigenvalue is supposed to be Fr{\'e}chet distribution \cite{auffinger2009poisson,soshnikov2004poisson}. Therefore, it is easy to consider that prior distribution can be set as Power-law or Fr{\'e}chet distribution when we introduce prior knowledge of the global spectrum or maximum eigenvalue. \par
Now we introduce the heavy-tailed prior upon the model. Firstly we consider the prior of global spectra of weight matrices. When the weight matrices are heavy-tailed, the LSD is power-law, so the prior distribution can be set as
\begin{equation}
    P\left( \mathbf{W} \right) =\prod_{l=1}^L{P\left( \mathbf{W}_l \right)}\propto \prod_{l=1}^L{\prod_{j=1}^{K_l}{\lambda _{l,j}^{-\alpha_l}}},
\end{equation}
where $\alpha_l$ is the tail index of the power-law of square singular value of weight matrix $\mathbf{W}_l$ in the $l$-th layer of the neural network, $\lambda_{l,j}$ is the $j$-th square singular value of $\mathbf{W}_l$. $K_l$ is the number of singular values of $\mathbf{W}_l$ that is considered to be from a heavy-tailed distribution. $K_l$ is a  hyperparameter and we choose $K_l$ as half the size of $\mathbf{W}_l$ in our study. Substituting this into (\ref{eq:bayes}),  we have the following optimization problem:
\begin{align}
\mathbf{W}^{\mathrm{MAP}}&=\underset{\mathbf{W}}{\mathrm{arg}\max}\sum_{i=1}^N{y_i\log f_{\mathbf{W}}\left( x_i \right)}-\sum_{l=1}^L{\sum_{j=1}^{K_l}{\alpha_l \log \lambda _{l,j}}} \label{eq:awr1}
\end{align}
Secondly, we consider the prior of the maximum square singular value of the weight matrices. When the weight matrices are heavy-tailed, the distribution of maximum square singular value is Fr{\'e}chet distribution, so the prior distribution can be set as
\begin{equation}
    P\left( \mathbf{W} \right) =\prod_{l=1}^L{P\left( \mathbf{W}_l \right)}=\prod_{l=1}^L{\exp \left( -\lambda _{\max,l}^{-\alpha_l} \right)}.
\end{equation}
where  $\alpha_l$ is the tail index of $\mathbf{W}_l$ and $\lambda_{\max,l}$ is the maximum square singular value. Similarly, substituting this into (\ref{eq:bayes}), we have the following optimization problem:
\begin{align}
\mathbf{W}^{\mathrm{MAP}}&=\underset{\mathbf{W}}{\mathrm{arg}\max}\sum_{i=1}^N{y_i\log f_{\mathbf{W}}\left( x_i \right)}-\sum_{l=1}^L{\lambda _{\max,l}^{-\alpha_l}}. \label{eq:awr2}
\end{align}

So far we derive two forms of MAP but we have some problems to solve: How to determine the hyperparameters $\boldsymbol{\alpha}=\{\alpha_l,1\leq l\leq L\}$, and how to solve this maximization problem. In Empirical Bayes, the hyperparameter is determined by maximizing the marginal likelihood of the data, that is
\begin{equation}
    \boldsymbol{\alpha} =\underset{\boldsymbol{\alpha}}{\mathrm{arg}\max}\log \int{P\left( \mathbf{y},\mathbf{W}\left| \mathbf{x},\boldsymbol{\alpha} \right. \right) \mathrm{d}\mathbf{W}}.
\end{equation}
It's apparently impossible since the integral is intractable. According to Mandt et al., \cite{mandt2017stochastic}, the SGD can be seen as a variational expectation maximization (VEM) method for Bayesian inference. The maximization problems in (\ref{eq:awr1}) and (\ref{eq:awr2}) are equivalent to the following minimization problem:
\begin{equation}
\label{eq:awr11}
\min_{\mathbf{W}} \quad \mathcal{L} \left( x,y \right) =\frac{1}{N}\sum_{i=1}^N{l\left( f\left( x_i \right) ,y_i \right)}+\sum_{l=1}^L{\sum_{j=1}^{K_l}{\alpha_l \log \lambda _{l,j}}},
\end{equation}
\begin{equation}
\label{eq:awr21}
\min_{\mathbf{W}} \quad \mathcal{L} \left( x,y \right) =\frac{1}{N}\sum_{i=1}^N{l\left( f\left( x_i \right) ,y_i \right)}+\sum_{l=1}^L{\lambda _{\max,l}^{-\alpha_l}},
\end{equation}
where $l(f(x),y)=-y\log f(x)$ is the cross entropy loss. The hyperparameters $\boldsymbol{\alpha}$ can be optimized when the SGD is seen as a type of VEM algorithm. Instead of the MLE, the Hill estimator is a better choice to estimate the hyperparameters $\boldsymbol{\alpha}$. When the tuning parameter is added, the (\ref{eq:awr11}) and (\ref{eq:awr21}) can be modified as:
\begin{equation}
\label{eq:awr12}
\min_{\mathbf{W}} \quad \mathcal{L} \left( x,y \right) =\frac{1}{N}\sum_{i=1}^N{l\left( f\left( x_i \right) ,y_i \right)}+\mu\sum_{l=1}^L{\sum_{j=1}^{K_l}{\hat{\alpha}_l \log \lambda _{l,j}}},
\end{equation}
\begin{equation}
\label{eq:awr22}
\min_{\mathbf{W}} \quad \mathcal{L} \left( x,y \right) =\frac{1}{N}\sum_{i=1}^N{l\left( f\left( x_i \right) ,y_i \right)}+\mu\sum_{l=1}^L{\lambda _{\max,l}^{-\hat{\alpha}_l}},
\end{equation}
where $\hat{\alpha}_l$ is the Hill estimator of the $l$-th layer weight matrix.\par
Note that (\ref{eq:awr12}) and (\ref{eq:awr22}) are the special cases of (\ref{prmp}) when $p_l(\mathbf W)=\alpha_l\cdot \sum_j \log \lambda_{j,l}$ and $p_l(\mathbf W)=\lambda_{\max,l}^{\alpha_l}$. Since the penalty terms are similar to the weighted alpha, these regularization terms can be considered as variants of Weighted Alpha. According to the priors used in these regularizations, we call the (\ref{eq:awr12}) \textbf{Heavy-Tailed Regularization under Power-law Prior} and the (\ref{eq:awr22}) \textbf{Heavy-Tailed Regularization under Fr{\'e}chet Prior}.

\section{Experiment}

In this section, we experimentally demonstrate the performance of Heavy-Tailed Regularization on multi-classification tasks. To verify the effectiveness of heavy-tailed regularization, we employed  the Weighted Alpha Regularization and Stable Rank Regularization with their variants, and the heavy-tailed regularization under the Power-law and Fr{\'e}chet spectral prior. Here all the tail index is replaced by its Hill estimator with $k=\frac{n}{2}$, where $n$ is the size of corresponding weight matrix.  In our experiment, we used the following training methods to compare with the heavy-tailed regularization approach:
\begin{enumerate}
    \item Vanilla problem (Base): We considered the original model without any explicit regularization.
    \item Weight Decay: We considered the most commonly used explicit regularization in (\ref{prmp}) where $p_l(\mathbf{W}) = \frac{1}{2} ||\mathbf W ||_F^2$.
    \item Spectral Norm Regularization: We considered another explicit regularization method with regards to the spectrum which penalty function $p_l(\mathbf{W}) = \frac{1}{2} ||\mathbf W ||_2^2$.
\end{enumerate}
All the experiments here are based on mini-batch SGD and learning rate decay. In our experiments, we used the following four settings on the model and dataset:
\begin{enumerate}
    \item The Three Layer Neural Network (FC3) on KMNIST and CIFAR10.
    \item The LeNet5 on CIFAR10.
    \item The ResNet18 on CIFAR10.
\end{enumerate}
\begin{figure}[htbp]
\centering
	\begin{minipage}{0.475\linewidth}
		\centering
    \includegraphics[width=0.9\linewidth]{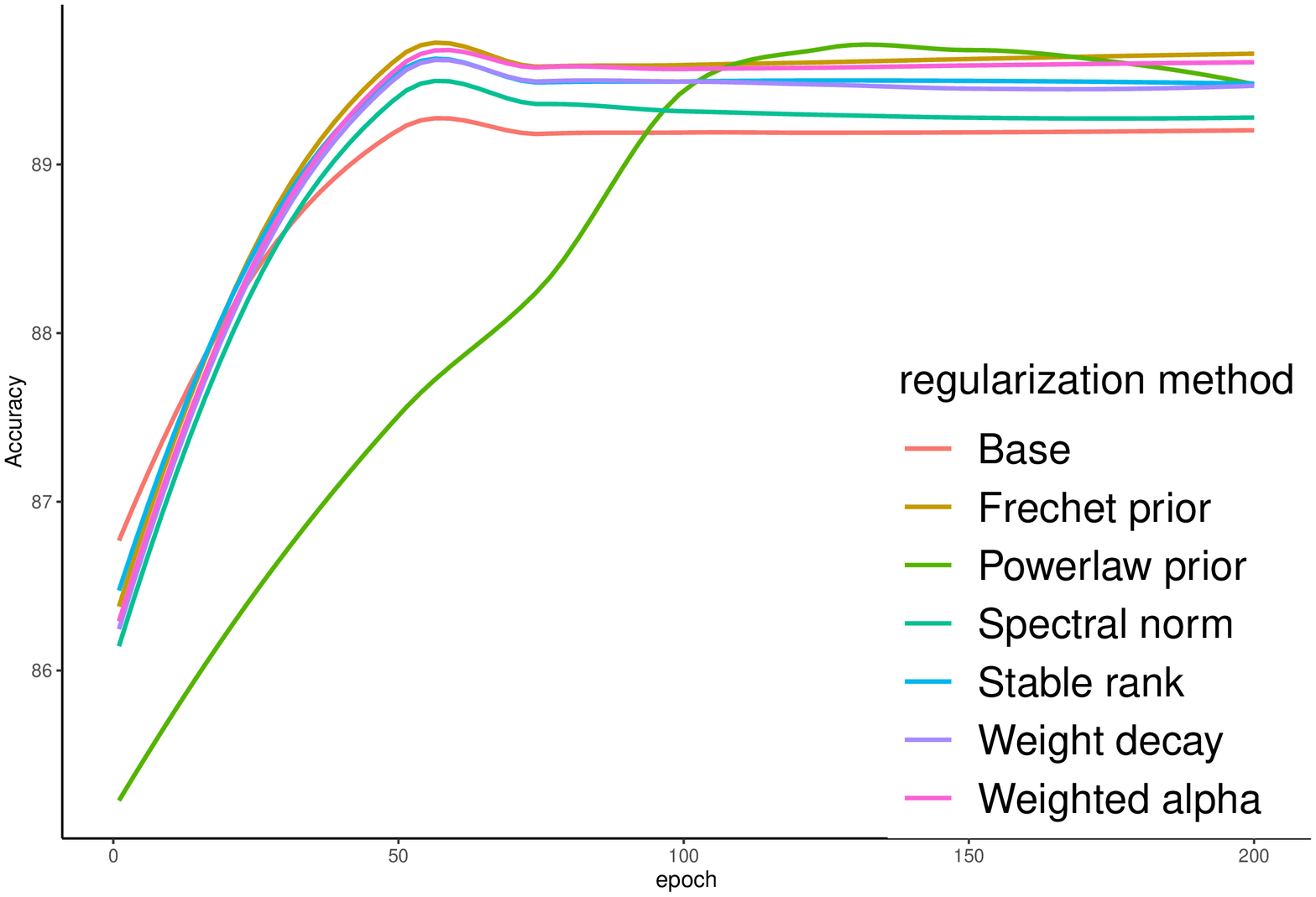}
    \caption{FC3 on KMNIST}
    \label{fig:nn3kmn}
	\end{minipage}
	\begin{minipage}{0.475\linewidth}
		\centering
    \includegraphics[width=0.9\linewidth]{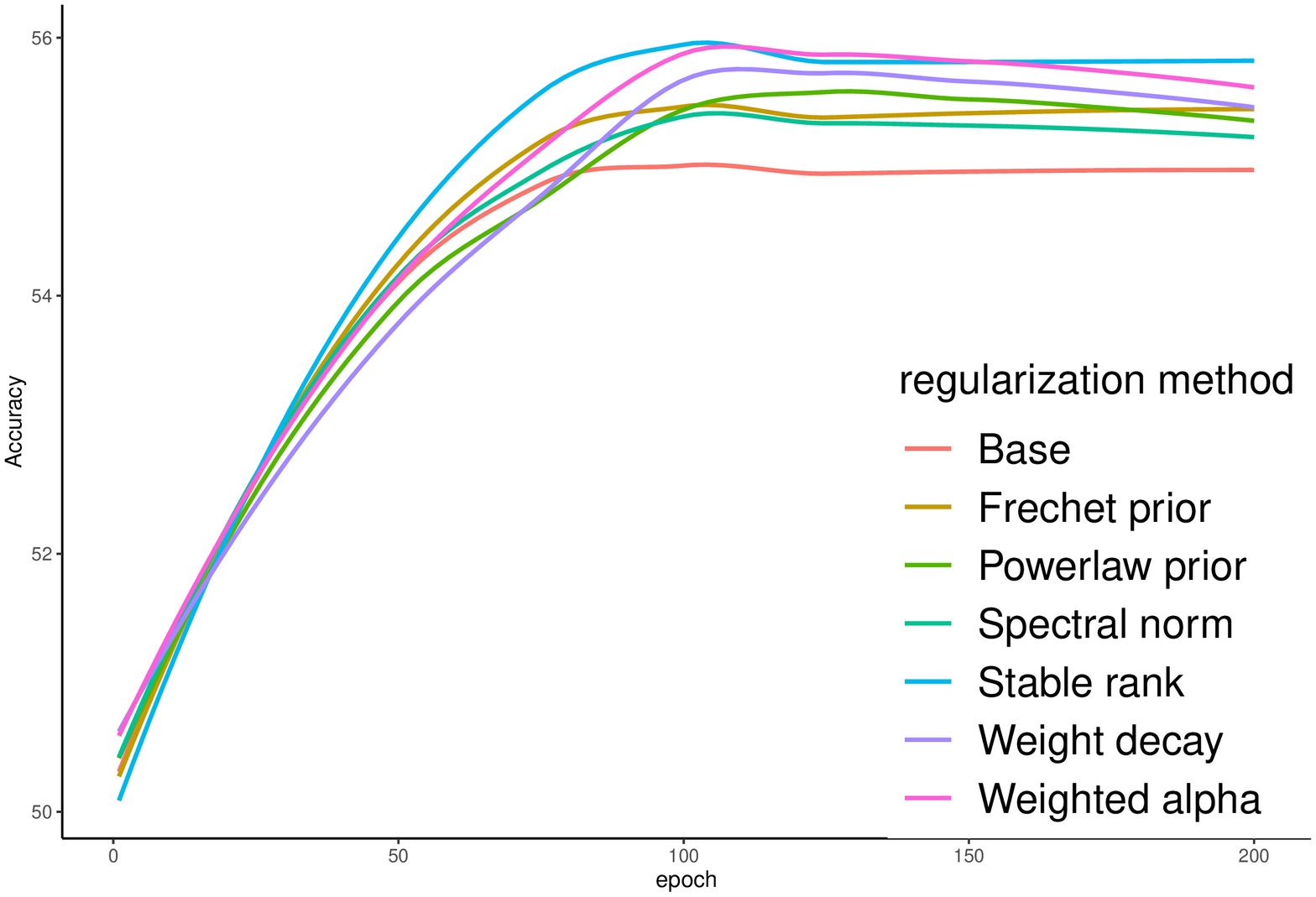}
    \caption{FC3 on CIFAR10}
    \label{fig:nn3cif}
	\end{minipage}
\end{figure}
\begin{figure}[t]
    \begin{minipage}{0.475\linewidth}
		\centering
    \includegraphics[width=0.9\linewidth]{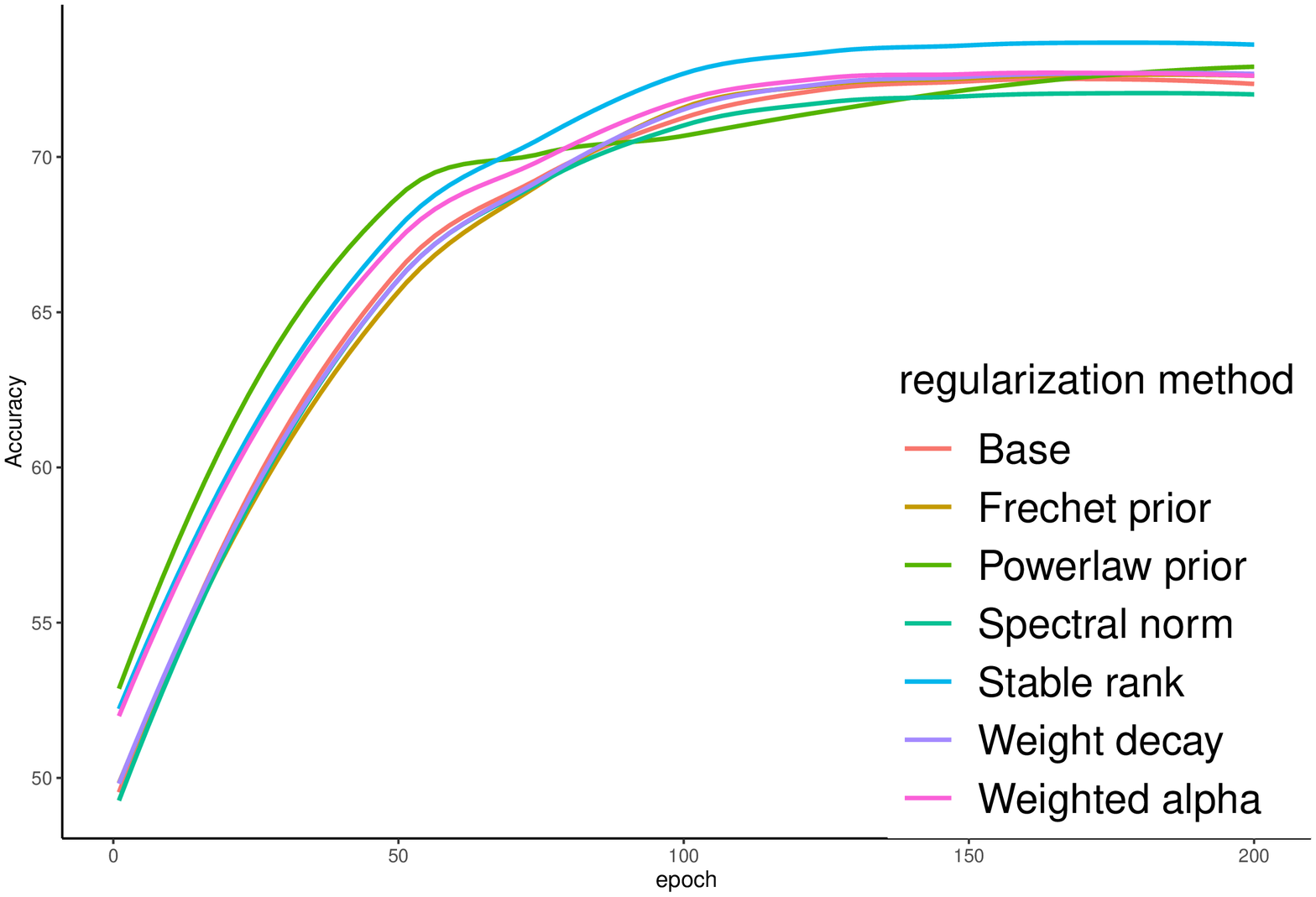}
    \caption{LeNet5 on CIFAR10}
    \label{fig:lecif}
	\end{minipage}
	\begin{minipage}{0.475\linewidth}
		\centering
    \includegraphics[width=0.9\linewidth]{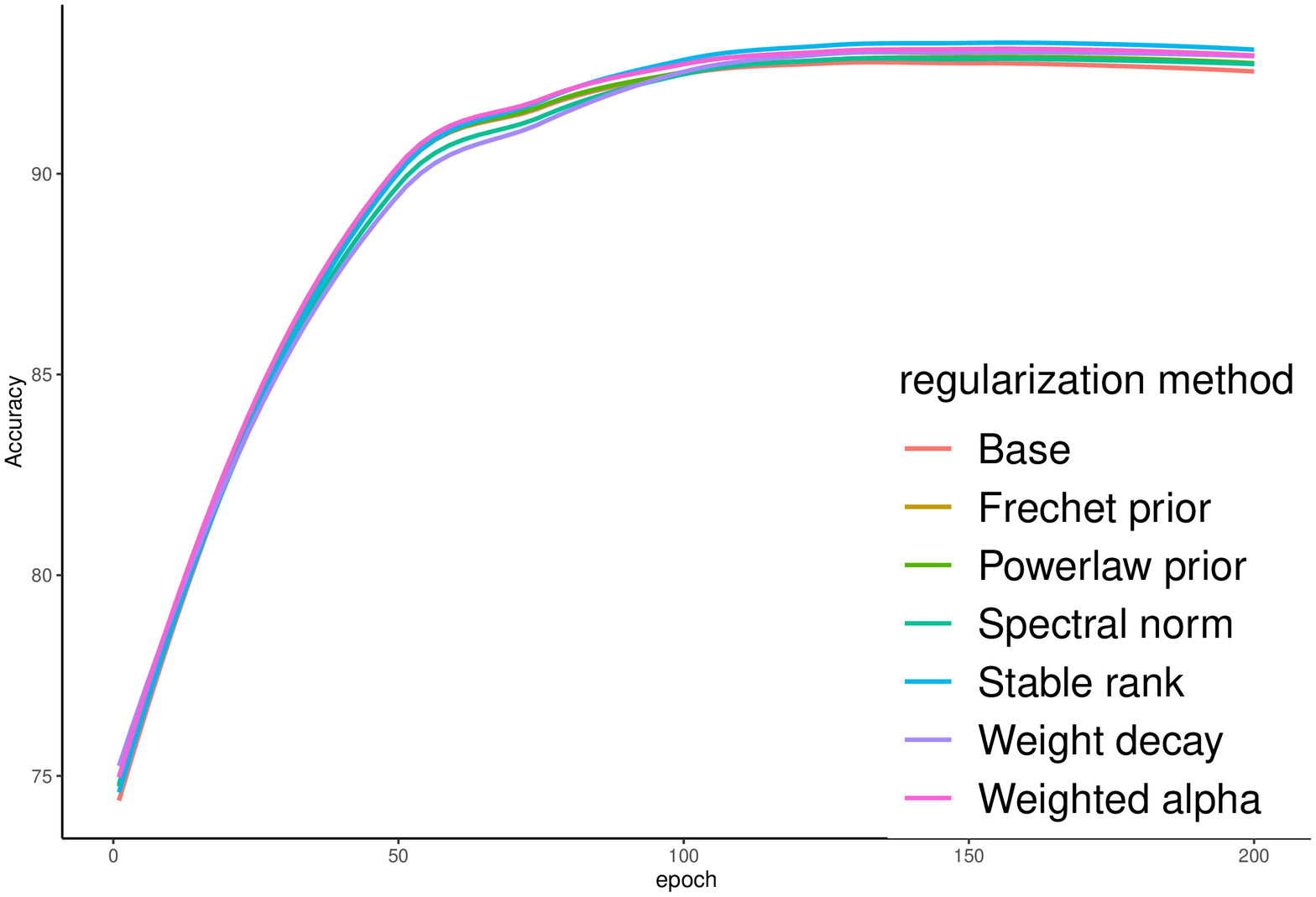}
    \caption{ResNet18 on CIFAR10}
    \label{fig:rescif}
	\end{minipage}
\end{figure}

\subsection{FC3}
In the first place, we train the neural network with three hidden layers on the KMNIST and CIFAR10 dataset for 200 epochs. The KMNIST dataset is adapted from Kuzushiji Dataset and is a drop-in replacement for the MNIST dataset. The image size of the KMNIST dataset is 28$\times$28. The CIFAR10 dataset consists of 60000 color images whose size is 32$\times$32. The CIFAR10 dataset is more complex than the KMNIST dataset so the CIFAR10 dataset is more difficult to be classified correctly. Because the image size of these two datasets varied, we use two types of three-layer neural networks with different sizes for each dataset. For the KMNIST dataset, we use the network with sizes of layers $\mathbf n = \left[784, 128, 128, 128, 10\right]$; For the CIFAR10 dataset, we use the network with sizes of layers $\mathbf n = \left[3072, 512, 256, 256, 10\right]$.
\par
The results are shown in Figure \ref{fig:nn3kmn}, \ref{fig:nn3cif} and Table \ref{tab:nn3}. The heavy-tailed regularizations all show better accuracies than the vanilla problem both on the KMNIST dataset and the CIFAR10 dataset. The Fr{\'e}chet prior achieves the best test accuracy on the KMNIST dataset, and the stable rank with lower threshold of $t=15$ achieves the best test accuracy on the CIFAR10 dataset.

\begin{table}[t]
\centering
\caption{The average (± standard error) of test accuracy of FC3 with different regularization methods on the KMNIST and CIFAR10 dataset}
\label{tab:nn3}
\resizebox{0.75\columnwidth}{!}{%
\begin{tabular}{@{}ccccc@{}}
\toprule
Network &
  Dataset &
  Method &
  $\beta$ &
  Test accuracy \\ \midrule
\multirow{14}{*}{FC3} &
  \multirow{7}{*}{KMNIST} &
  base &
   &
  89.19±0.020 \\
 &
   &
  weight decay &
  5.00$\times 10^{-4}$ &
  89.44±0.037 \\
 &
   &
  spectral norm &
  0.0001 &
  89.27±0.013 \\
 &
   &
  weighted alpha\tablefootnote[1]{We use power decay weighted alpha with $k=2$.\label{ft:pw}} &
  5.00$\times 10^{-5}$ &
  89.60±0.011 \\
 &
   &
  stable rank\tablefootnote[2]{We use lower threshold stable rank with $t=1$.\label{ft:ls}} &
  1.00$\times 10^{-4}$ &
  89.48±0.008 \\
 &
   &
  Power-law prior &
  5.00$\times 10^{-4}$ &
  89.58±0.175 \\
 &
   &
  \textbf{Fr{\'e}chet prior} &
  \textbf{2.00$\times 10^{-5}$} &
  \textbf{89.64±0.173} \\ \cmidrule(l){2-5} 
 &
  \multirow{7}{*}{CIFAR10} &
  base &
   &
  54.97±0.039 \\
 &
   &
  weight decay &
  5.00$\times 10^{-4}$ &
  55.56±0.092 \\
 &
   &
  spectral norm &
  0.0001 &
  55.27±0.003 \\
 &
   &
  weighted alpha\tablefootnotemark{ft:pw} &
  1.00$\times 10^{-4}$ &
  55.72±0.053 \\
 &  & \textbf{stable rank}\tablefootnotemark{ft:ls} & \textbf{1.00$\times 10^{-4}$} & \textbf{55.82±0.041} \\
 &
   &
  Power-law prior &
  5.00$\times 10^{-5}$ &
  55.44±0.038 \\
 &
   &
  Fr{\'e}chet prior &
  5.00$\times 10^{-5}$ &
  55.45±0.029 \\ \bottomrule
\end{tabular}%
}
\end{table}
\subsection{LeNet5}
Secondly, we train the LeNet5 on the CIFAR10 dataset for 200 epochs. The LeNet5 is a famous and classical convolutional neural network (CNN) architecture. The results are shown in Figure \ref{fig:lecif} and Table \ref{tab:lecif}. The heavy-tailed regularization also all shows better accuracy than the vanilla problem both on the CIFAR10 dataset. As shown in the table, the stable rank with $\beta=0.1$ achieves the best test accuracy.
\begin{table}[t]
\centering
\caption{The average (± standard error) of test accuracy of LeNet5 with different regularization methods on the CIFAR10 dataset}
\label{tab:lecif}
\resizebox{0.75\columnwidth}{!}{%
\begin{tabular}{@{}ccccc@{}}
\toprule
Network                 & Dataset                  & Method               & $\beta$             & Test Accuracy        \\ \midrule
\multirow{7}{*}{LeNet5} & \multirow{7}{*}{CIFAR10} & base                 &                     & 72.42±0.213          \\
                        &                          & weight decay         & 5.00$\times10^{-4}$ & 72.62±0.277          \\
                        &                          & spectral norm        & 0.0001              & 71.98±0.275          \\
                        &                          & weighted alpha\tablefootnotemark{ft:pw}       & 0.004               & 72.61±0.300          \\
                        &                          & \textbf{stable rank} & \textbf{0.1}        & \textbf{73.63±0.193} \\
                        &                          & Power-law prior       & 7.00$\times10^{-4}$ & 72.61±1.061          \\
                        &                          & Frechet prior        & 5.00$\times10^{-5}$ & 72.58±0.270          \\ \bottomrule
\end{tabular}%
}
\end{table}

\subsection{ResNet18}
Thirdly, we train the ResNet18 on the CIFAR10 dataset for 200 epochs. The ResNet is a CNN architecture which greatly advanced the SOTA in various computer vision tasks. In this experiment, we add one linear layer with size of 512$\times$128 before the linear layer in the origin ResNet18 architecture. The results are shown in Figure \ref{fig:rescif} and Table \ref{tab:resnetcif}. As shown in the table, the stable rank with $\beta=5\times10^{-4}$ achieves the best test accuracy.
\begin{table}[t]
\centering
\caption{The average (± standard error) of test accuracy of ResNet18 with different regularization methods on the CIFAR10 dataset}
\label{tab:resnetcif}
\resizebox{0.75\columnwidth}{!}{%
\begin{tabular}{@{}ccccc@{}}
\toprule
Network                   & Dataset                  & Method               & $\beta$                      & Test accuracy        \\ \midrule
\multirow{7}{*}{ResNet18} & \multirow{7}{*}{CIFAR10} & base                 &                              & 92.65±0.066          \\
                          &                          & weight decay         & 5.00$\times10^{-4}$          & 93.15±0.087          \\
                          &                          & spectral norm        & 0.0001                       & 92.78±0.069          \\
                          &                          & weighted alpha       & 5.00$\times10^{-5}$          & 93.04±0.045          \\
                          &                          & \textbf{stable rank} & \textbf{5.00$\times10^{-4}$} & \textbf{93.19±0.049} \\
                          &                          & Power-law prior       & 1.00$\times10^{-4}$          & 92.85±0.111          \\
                          &                          & Frechet prior        & 3.00$\times10^{-5}$          & 93.07±0.086          \\ \bottomrule
\end{tabular}%
}
\end{table}

%
%
%
\bibliographystyle{splncs04}
\bibliography{ref}

\begin{thebibliography}{10}
\providecommand{\url}[1]{\texttt{#1}}
\providecommand{\urlprefix}{URL }
\providecommand{\doi}[1]{https://doi.org/#1}

\bibitem{auffinger2009poisson}
Auffinger, A., Ben~Arous, G., P{\'e}ch{\'e}, S.: Poisson convergence for the
  largest eigenvalues of heavy tailed random matrices. In: Annales de l'IHP
  Probabilit{\'e}s et statistiques. vol.~45, pp. 589--610 (2009)

\bibitem{barsbey2021heavy}
Barsbey, M., Sefidgaran, M., Erdogdu, M.A., Richard, G., Simsekli, U.: Heavy
  tails in sgd and compressibility of overparametrized neural networks.
  Advances in Neural Information Processing Systems  \textbf{34},  29364--29378
  (2021)

\bibitem{bartlett1998almost}
Bartlett, P., Maiorov, V., Meir, R.: Almost linear vc dimension bounds for
  piecewise polynomial networks. Advances in neural information processing
  systems  \textbf{11} (1998)

\bibitem{bartlett2017spectrally}
Bartlett, P.L., Foster, D.J., Telgarsky, M.J.: Spectrally-normalized margin
  bounds for neural networks. Advances in neural information processing systems
   \textbf{30} (2017)

\bibitem{bartlett2002rademacher}
Bartlett, P.L., Mendelson, S.: Rademacher and gaussian complexities: Risk
  bounds and structural results. Journal of Machine Learning Research
  \textbf{3}(Nov),  463--482 (2002)

\bibitem{Chen_Zhao_Li_Huang_Ou_2019}
Chen, Q., Zhao, H., Li, W., Huang, P., Ou, W.: Behavior sequence transformer
  for e-commerce recommendation in alibaba. In: Proceedings of the 1st
  International Workshop on Deep Learning Practice for High-Dimensional Sparse
  Data (Dec 2019). \doi{10.1145/3326937.3341261},
  \url{http://dx.doi.org/10.1145/3326937.3341261}

\bibitem{davis2016extreme}
Davis, R.A., Heiny, J., Mikosch, T., Xie, X.: Extreme value analysis for the
  sample autocovariance matrices of heavy-tailed multivariate time series.
  Extremes  \textbf{19}(3),  517--547 (2016)

\bibitem{davis2016asymptotic}
Davis, R.A., Mikosch, T., Pfaffel, O.: Asymptotic theory for the sample
  covariance matrix of a heavy-tailed multivariate time series. Stochastic
  Processes and their Applications  \textbf{126}(3),  767--799 (2016)

\bibitem{davis2014limit}
Davis, R.A., Pfaffel, O., Stelzer, R.: Limit theory for the largest eigenvalues
  of sample covariance matrices with heavy-tails. Stochastic Processes and
  their Applications  \textbf{124}(1),  18--50 (2014)

\bibitem{Galassi_Lippi_Torroni_2020}
Galassi, A., Lippi, M., Torroni, P.: Attention in natural language processing.
  IEEE Transactions on Neural Networks and Learning Systems  \textbf{32}(10),
  4291–4308 (Sep 2020). \doi{10.1109/tnnls.2020.3019893},
  \url{http://dx.doi.org/10.1109/tnnls.2020.3019893}

\bibitem{he2016deep}
He, K., Zhang, X., Ren, S., Sun, J.: Deep residual learning for image
  recognition. In: Proceedings of the IEEE conference on computer vision and
  pattern recognition. pp. 770--778 (2016)

\bibitem{hodgkinson2021multiplicative}
Hodgkinson, L., Mahoney, M.: Multiplicative noise and heavy tails in stochastic
  optimization. In: International Conference on Machine Learning. pp.
  4262--4274. PMLR (2021)

\bibitem{mandt2017stochastic}
Mandt, S., Hoffman, M.D., Blei, D.M.: Stochastic gradient descent as
  approximate bayesian inference. arXiv preprint arXiv:1704.04289  (2017)

\bibitem{martin2019traditional}
Martin, C.H., Mahoney, M.W.: Traditional and heavy-tailed self regularization
  in neural network models. arXiv preprint arXiv:1901.08276  (2019)

\bibitem{martin2020heavy}
Martin, C.H., Mahoney, M.W.: Heavy-tailed universality predicts trends in test
  accuracies for very large pre-trained deep neural networks. In: Proceedings
  of the 2020 SIAM International Conference on Data Mining. pp. 505--513. SIAM
  (2020)

\bibitem{martin2021implicit}
Martin, C.H., Mahoney, M.W.: Implicit self-regularization in deep neural
  networks: Evidence from random matrix theory and implications for learning.
  Journal of Machine Learning Research  \textbf{22}(165),  1--73 (2021)

\bibitem{martin2021post}
Martin, C.H., Mahoney, M.W.: Post-mortem on a deep learning contest: a
  simpson's paradox and the complementary roles of scale metrics versus shape
  metrics. arXiv preprint arXiv:2106.00734  (2021)

\bibitem{martin2021predicting}
Martin, C.H., Peng, T.S., Mahoney, M.W.: Predicting trends in the quality of
  state-of-the-art neural networks without access to training or testing data.
  Nature Communications  \textbf{12}(1),  1--13 (2021)

\bibitem{meng2021impact}
Meng, X., Yao, J.: Impact of classification difficulty on the weight matrices
  spectra in deep learning and application to early-stopping. arXiv preprint
  arXiv:2111.13331  (2021)

\bibitem{nagarajan2019uniform}
Nagarajan, V., Kolter, J.Z.: Uniform convergence may be unable to explain
  generalization in deep learning. Advances in Neural Information Processing
  Systems  \textbf{32} (2019)

\bibitem{neyshabur2017pac}
Neyshabur, B., Bhojanapalli, S., Srebro, N.: A pac-bayesian approach to
  spectrally-normalized margin bounds for neural networks. arXiv preprint
  arXiv:1707.09564  (2017)

\bibitem{simsekli2020hausdorff}
Simsekli, U., Sener, O., Deligiannidis, G., Erdogdu, M.A.: Hausdorff dimension,
  heavy tails, and generalization in neural networks. Advances in Neural
  Information Processing Systems  \textbf{33},  5138--5151 (2020)

\bibitem{soshnikov2004poisson}
Soshnikov, A.: Poisson statistics for the largest eigenvalues of wigner random
  matrices with heavy tails. Electronic Communications in Probability
  \textbf{9},  82--91 (2004)

\bibitem{vapnik1994measuring}
Vapnik, V., Levin, E., Le~Cun, Y.: Measuring the vc-dimension of a learning
  machine. Neural computation  \textbf{6}(5),  851--876 (1994)

\bibitem{Vaswani2017AttentionIA}
Vaswani, A., Shazeer, N.M., Parmar, N., Uszkoreit, J., Jones, L., Gomez, A.N.,
  Kaiser, L., Polosukhin, I.: Attention is all you need. ArXiv
  \textbf{abs/1706.03762} (2017)

\bibitem{zhang2021understanding}
Zhang, C., Bengio, S., Hardt, M., Recht, B., Vinyals, O.: Understanding deep
  learning (still) requires rethinking generalization. Communications of the
  ACM  \textbf{64}(3),  107--115 (2021)

\bibitem{zhou2020towards}
Zhou, P., Feng, J., Ma, C., Xiong, C., Hoi, S.C.H., et~al.: Towards
  theoretically understanding why sgd generalizes better than adam in deep
  learning. Advances in Neural Information Processing Systems  \textbf{33},
  21285--21296 (2020)

\end{thebibliography}

\end{document}